%% file: conference_101719.tex
\documentclass[conference]{IEEEtran}
\IEEEoverridecommandlockouts
\usepackage{cite}
\usepackage{amsmath,amssymb,amsfonts}
\usepackage{algorithmic}
\usepackage{graphicx}
\usepackage{textcomp}
\usepackage{xcolor}
\usepackage{multirow}
\usepackage{tabularx}
\usepackage{booktabs}
\usepackage{makecell}
\usepackage[ruled,linesnumbered]{algorithm2e}
\def\BibTeX{{\rm B\kern-.05em{\sc i\kern-.025em b}\kern-.08em
    T\kern-.1667em\lower.7ex\hbox{E}\kern-.125emX}}
    
\makeatletter
\newcommand{\linebreakand}{%
  \end{@IEEEauthorhalign}
  \hfill\mbox{}\par
  \mbox{}\hfill\begin{@IEEEauthorhalign}
}
\makeatother

\begin{document}


\title{Affective-NLI: Towards Accurate and Interpretable Personality Recognition in Conversation}



\author{
\IEEEauthorblockN{Zhiyuan Wen, Jiannong Cao, Yu Yang, Ruosong Yang, Shuaiqi Liu}
\IEEEauthorblockA{\textit{Department of Computing}, \textit{The Hong Kong Polytechnic University}, Hong Kong, China \\
\{cszwen, jiannong.cao, cs-yu.yang, rsong.yang, shuaiqi.liu\}@polyu.edu.hk}
}

\maketitle

\begin{abstract}
Personality Recognition in Conversation (PRC) aims to identify the personality traits of speakers through textual dialogue content. 
It is essential for providing personalized services in various applications of Human-Computer Interaction (HCI), such as AI-based mental therapy and companion robots for the elderly. 
Most recent studies analyze the dialog content for personality classification yet overlook two major concerns that hinder their performance. 
First, crucial implicit factors contained in conversation, such as emotions that reflect the speakers' personalities are ignored.
Second, only focusing on the input dialog content disregards the semantic understanding of personality itself, which reduces the interpretability of the results.
In this paper, we propose Affective Natural Language Inference (Affective-NLI) for accurate and interpretable PRC. 
To utilize affectivity within dialog content for accurate personality recognition, we fine-tuned a pre-trained language model specifically for emotion recognition in conversations, facilitating real-time affective annotations for utterances. 
For interpretability of recognition results, we formulate personality recognition as an NLI problem by determining whether the textual description of personality labels is entailed by the dialog content. Extensive experiments on two daily conversation datasets suggest that Affective-NLI significantly outperforms (by 6\%-7\%) state-of-the-art approaches. Additionally, our \textbf{Flow} experiment demonstrates that Affective-NLI can accurately recognize the speaker's personality in the early stages of conversations by surpassing state-of-the-art methods with 22\%-34\%\footnote{Our source code and data is at https://github.com/preke/Affective-NLI.}.
\end{abstract}


\input{intro_new.tex}
\input{related_works_new.tex}

\input{preliminary.tex}
\input{method_new.tex}

\input{experiment_new.tex}

\input{result_analysis_new.tex}

\input{limitation.tex}

\section{Conclusion, Limitations and Future Work}

This paper presents Affective-NLI, a simple yet effective approach for enhancing dialog content with affective information and personality label descriptions, enabling accurate personality recognition in conversations. Affective-NLI provides interpretable recognition results and is easy to implement with various architectures of pre-trained language models. Through extensive experiments, we validate the effectiveness of Affective-NLI in personality recognition. Furthermore, Affective-NLI demonstrates its suitability for early-stage personality recognition in conversations.

Although the effectiveness of Affective-NLI has been validated, we identify some limitations that can be improved and expanded upon in future work. First, Affective-NLI follows the setting in existing studies on personality recognition in conversation, where personality recognition is performed through independent binary classification of each trait. However, this setting overlooks the correlations among different personality traits, which have been explored in psychology findings \cite{mccrae1987validation, goldberg1993structure}. In future research, we will also investigate how to incorporate these correlations into Affective-NLI for more accurate personality recognition. Second, Affective-NLI focuses solely on the textual modality. Personality can be manifested in non-verbal factors such as facial expressions \cite{naumann2009personality} and face gestures \cite{lathaface}, or physiological signals like Electrocardiogram (ECG) and heart rate\cite{shui2023personality}. In future work, we will also investigate transforming signals collected from pervasive systems into textual descriptions and achieving multimodal personality recognition.

\section*{Acknowledgement}
This work is conducted at the Research Institute for Artificial Intelligence of Things (RIAIoT) at PolyU and supported by the Hong Kong Jockey Club Charities Trust fund (No. 2021-0369), and PolyU Research and Innovation Office (No. BD4A).

\bibliographystyle{IEEEtran}
\bibliography{IEEEabrv,mybibfile}

\end{document}

%% file: intro_new.tex
\section{Introduction}

\begin{figure}[t]
\centering
\includegraphics[scale=0.5]{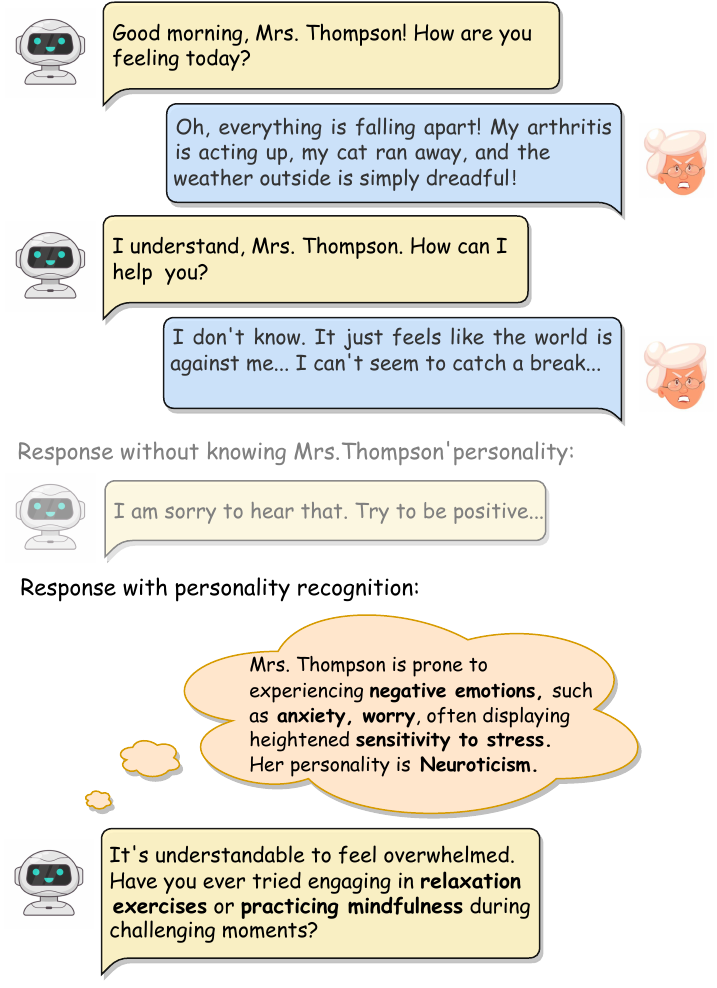}
\caption{When the robot lacks knowledge about Mrs. Thompson's personality, its comfort is generic and straightforward. However, if the robot deduces that Mrs. Thompson exhibits \textit{Neuroticism} based on the dialog content, it can tailor its suggestions to offer more personalized and comforting advice.}
\label{toy_example}
\end{figure}

The advancement of large language models (\textit{e.g.}, ChatGPT) has led to the emergence of numerous applications within Human-Computer Interaction (HCI), including AI-based mental therapy \cite{d2020ai}, educational robots \cite{yang2020hybrid}, and companion robots designed for the elderly \cite{kahambing2023chatgpt}. In these application scenarios, fast and accurately recognizing the user's personality from the dialog content is crucial, as it enables a better understanding of the user's personalized needs and requirements, ultimately leading to the provision of high-quality services \cite{chang2018stereotypes,srinarong2021development,fatahi2016survey}, as shown in Figure \ref{toy_example}.


Although important, addressing personality recognition in conversation (PRC) also faces a significant challenge. PRC strives to classify the specified speaker's personality into pre-defined personality classes (\textit{e.g.}, the big five personalities \cite{costa1992normal}  shown in Table \ref{intro ocean traits}) by analyzing the dialog content. However, personalities are relatively permanent traits that describe the consistency and individuality of a person's behaviors \cite{feist2012j}. Accurate personality assessments typically involve obtaining detailed personal information or long-term historical data, such as self-report essays, questionnaires, or longitudinal records from social media, which are always unavailable in daily conversation scenarios. 

Recent studies solve the PRC problem utilizing the advanced semantic comprehension abilities of deep neural networks to analyze the dialog content. Specifically, researchers either design dialog encoding models \cite{rissola2019personality} or fine-tune pre-trained language models \cite{jiang2020automatic, chen2022cped, wen2023desprompt} for PRC. However, these methods primarily concentrate on the explicit dialog content yet overlook two major concerns that hinder their recognition performance. First, semantic information contained in short utterances within conversations is inherently limited. These methods ignore crucial implicit factors contained in conversation, such as emotions and attitudes that reflect the speakers' personalities, so that hinders their recognition accuracies. Second, only focusing on the input dialog content disregards the semantic understanding of personality class labels, which loses sight of additional discriminative cues and reduces the interpretability of the results.

\begin{table}[t]
    \centering
	\small
	\renewcommand\arraystretch{1.3}
    \caption{The big five personality traits (OCEAN) and description.}
    \begin{tabular}{ll}
        \hline  
		\textbf{Traits} & \textbf{Descriptions}\\
        \hline  
		Openness & Openminded, imaginative, and sensitive. \\		
		Conscientiousness & Scrupulous, well-organized.\\
		Extraversion & Tend to experience positive emotions. \\
		Agreeableness & Trusting, sympathetic, and cooperative. \\
		Neuroticism	&  Tend to experience psychological distress. \\
		\hline
    \end{tabular}
    \label{intro ocean traits}
\end{table}

To fill in the research gap, we propose Affective-Natural Language Inference (Affective-NLI), aiming for accurate and interpretable personality recognition in conversation with affective annotations (\textit{i.e.}, discrete emotion category labels) of dialog content and the natural language descriptions of personality labels. Psychology findings suggest that there are strong correlations between personalities and affective expressions in conversations \cite{mehrabian1995relationships,mehrabian1996analysis}. However, accurately obtaining real-time affective annotations of dialog content for personality recognition is not a trivial task. We fine-tuned a pre-trained language model specifically for emotion recognition in conversations, facilitating real-time affective annotations for the utterances. Besides, the semantics behind personality labels contain descriptions of individual behavioral characteristics and tendencies in expressing affectivities, offering additional grounds for personality recognition.
To utilize them for our problem, we adopt text descriptions of personality labels from psychology findings \cite{saucier1996evidence,roccas2002big} and formulate personality recognition as a natural language inference (NLI) problem. Specifically, the objective is to determine whether the personality description of the speaker (as the hypothesis) is correct or not, given the dialog content (as the premise). We solve this NLI problem by constructing a natural language prompt, including the premise and the hypothesis above, to fine-tune the pre-trained language models.


Compared to existing approaches, Affective-NLI leverages auxiliary affective information to enhance the limited semantics in dialog content. Then, Affective-NLI correlates the language and behavior of speakers within conversations with descriptions of personality traits, yielding more interpretable recognition results. Moreover, Affective-NLI mainly manipulates the input dialog content, making it compatible with large pre-trained language models in various architectures.

We conducted extensive experiments on two daily conversation datasets annotated with personality labels. In the standard PRC setting, where the full dialog input is provided, Affective-NLI demonstrated a significant improvement (6\%-7\%) in the accuracy of personality recognition compared to state-of-the-art methods. Additionally, we designed a \textbf{Flow} experiment to assess whether Affective-NLI can instantly recognize the speaker's personality during the early stages of a conversation. The results showed that with just one or two utterances from the speaker under analysis, Affective-NLI achieved an accuracy of around 0.5-0.6 in personality recognition, surpassing state-of-the-art methods by 22\%-34\%. These findings suggest that Affective-NLI can be effectively employed for personality recognition in real-world human-computer interaction (HCI) applications. To demonstrate the interpretability of Affective-NLI, we conduct a case study to illustrate how our methods obtain the personality recognition results in a conversation example.

To summarize, our contributions are listed as follows:

\begin{itemize}
\item We introduce Affective-Natural Language Inference (Affective-NLI) to facilitate personality recognition in conversation with affective annotations of dialog content and the text descriptions of personality labels.
\item Affective-NLI is capable of providing interpretable recognition results by correlating the language and behavior of speakers within conversations with descriptions of personality traits. 
\item Extensive experiments conducted on two conversational datasets verify the effectiveness of Affective-NLI. Additionally, Affective-NLI has been validated for early-stage personality recognition during conversations.
\end{itemize}

%% file: related_works_new.tex
\section{Related Works}

\begin{figure*}[t]
\centering
\includegraphics[scale=0.5]{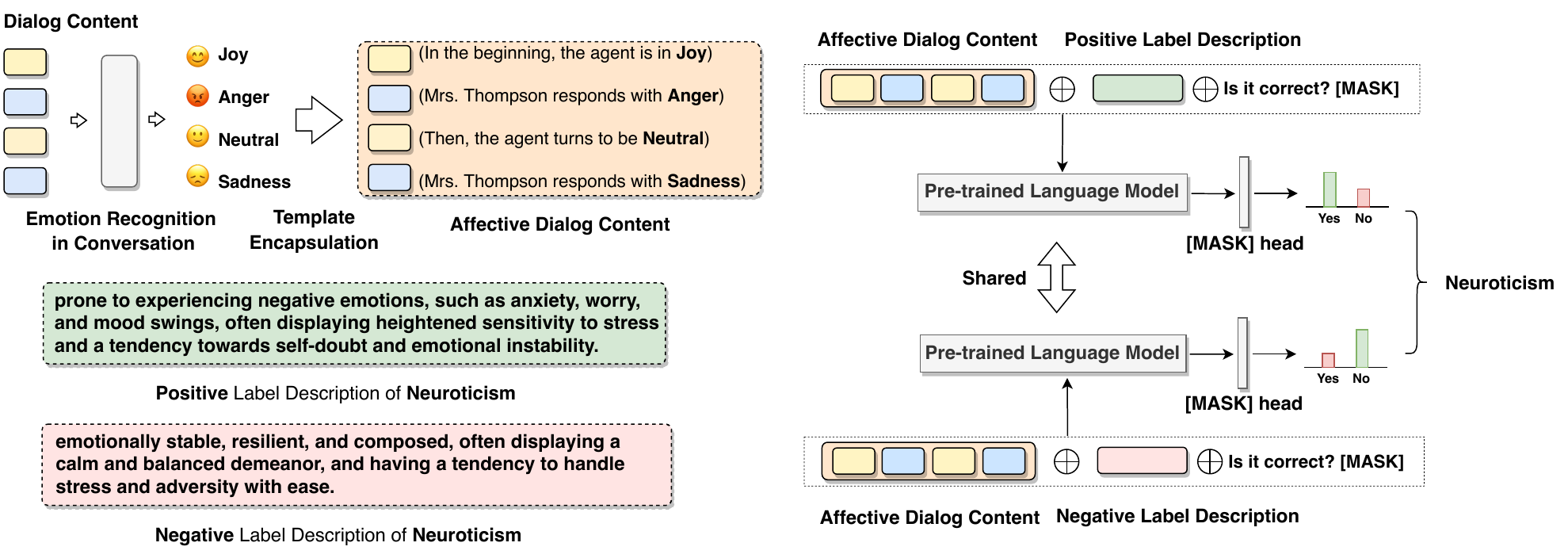}
\caption{An overview of Affective-NLI with the first four utterances in Figure \ref{toy_example}. The upper-left part illustrates the affective dialog content construction. The lower-left part shows the positive and negative label descriptions of \textit{Neuroticism}. The right part shows (1) how we construct NLI prompting samples with the affective dialog content and both positive and negative personality label descriptions and (2) how we combine both NLI results for the final personality recognition. Noting that the two different inputs to the same pre-trained language model.}
\label{model}
\end{figure*}

In this section, we review existing research related to personality recognition in conversation. 

Accurately recognizing users' personalities in conversation is crucial for various applications, like building user trust in conversational agents \cite{bickmore2001relational}, providing considerate elderly companions \cite{chang2018stereotypes}, or delivering personalized customer service\cite{srinarong2021development}. So, this task has long been a critical research issue and explored by many researchers.

Research in early stages focus on finding statistical features for recognizing personality. The statistical word usage patterns and social behavior habits are highly correlated to the personality traits \cite{pennebaker2001linguistic}. Informed by these findings, \cite{mehl2006personality} tracked 96 participants using the Electronically Activated Recorder (EAR) to examine the expression of personality in daily conversations. Besides, \cite{mairesse2006automatic} firstly presented several non-linear statistical models to rank utterances on the Big Five personality traits with linguistic features. However, although the shallow features are efficient in providing statistical differences for personality recognition, they fail when the personalities are identified by deep understanding the dialog content.

With the development of deep learning, neural network models are widely applied to recognizing personality through understanding the semantics of the conversation. \cite{rissola2019personality} proposed a classification model based on capsule neural networks to extract meaningful hidden patterns from conversations and use them to assess the personality of individuals. Subsequently, \cite{jiang2020automatic} used the pre-trained language models to encode the dialog content in different forms for personality classification. In \cite{chen2022cped}, the authors inputted the full conversation text into BERT \cite{devlin2018bert} and five independent SeNets \cite{hu2018squeeze} as the feature fusion layer of local personality features of different conversations. Some researchers also investigate prompt-tuning pre-trained language models (PLMs) for personality recognition. DesPrompt \cite{wen2023desprompt} utilizes a personality-descriptive prompt to encapsulate the input content, hence reformulates personality recognition as a cloze-style word-filling task to align with the pre-training Masked Language Modeling (MLM) task in PLMs.

However, in conversation scenarios, the utterances used for personality recognition are short sentences with limited quantity. Only focusing on semantic information in utterances fails to provide sufficient information to analyze personality, which hinders the recognition accuracy. Besides, concentrating on the input dialog content neglects the semantic comprehension of personality itself, thereby overlooking valuable discriminative cues and diminishing the interpretability of the results.

%% file: method_new.tex
\section{Affective-NLI}

In this section, we introduce the proposed Affective-NLI (as shown in Figure \ref{model}). We begin by presenting the formulation of the PRC problem in Section \ref{pf}. Subsequently, we delve into the construction of affective dialog content by Affective-NLI in Section \ref{adc}, followed by the introduction of the personality label descriptions utilized in Section \ref{pd}. Lastly, we introduce how we construct NLI samples for fine-tuning and inference in pre-trained language models in Section \ref{ft}.

\subsection{Problem Formulation}
\label{pf}
We study the Personality Recognition in Conversation (PRC) problem as stated below: 

Given the conversation content $X = \{u_1, u_2, ... , u_m\}$ with $m$ text utterances $u_i$ among a speaker $s$ and other speakers, the objective is to recognize the personality traits of $s$, denoted as $y$. $y$ is represented as a 5-$d$ binary vector $[y_a, y_c, y_e, y_n,y_o]$ indicating the binary values of each trait referring to the big-five personality traits theory (shown in Table \ref{intro ocean traits}, shortened as \textbf{AGR}, \textbf{CON}, \textbf{EXT}, \textbf{NEU}, and \textbf{OPN}, respectively). The big-five trait theory offers a categorical classification of personality derived from both the trait theory and the lexical hypothesis in psychology. It is widely used in analysis for social media content \cite{iacobelli2011large,souri2018personality,ji2023chatgpt, rao2023chatgpt} and conversations \cite{mairesse2006automatic}.

We propose Affective-Natural Language Inference (Affective-NLI) to formulate the PRC problem as the NLI problem between the affective dialog content (\textit{premise}) and the descriptions of personality labels (\textit{hypothesis}). Natural Language Inference (NLI) determines whether a hypothesis is true (entailment), false (contradiction), or undetermined (neutral) given a premise. In the scenario of PRC, Affective-NLI determines whether the \textit{hypothesis} of the personality label description for the speaker $s$ is correct or not, given the affective dialog content as the \textit{premise}.

Formally, to recognize each specific personality trait $y_p \in [y_a, y_c, y_e, y_o, y_n]$ for the speaker $s$ given input dialog content, Affective-NLI maximizes the sum of the two probabilities:

 \begin{equation}
 	\small
	\begin{aligned}
P_{pos} + P_{neg} = P(y_p | X_{af}; H_{pos}(p)) + P(1-y_p | X_{af}; H_{neg}(p))
	\end{aligned}
	\label{prob}
\end{equation}
where the $X_{af} = \{(u_1, a_1), (u_2, a_2), ... , (u_m, a_m) \}$ is the affective dialog content obtained by attaching an affective annotation $a_i$ to each utterance $u_i$. $H_{pos}(p)$ and $H_{neg}(p)$ are the natural language descriptions of people with and without the specific personality trait $p$, respectively.

Note that $H_{pos}(p)$ and $H_{neg}(p)$ are opposite descriptions to a same personality trait $p$. The principle of maximizing both $P_{pos}$ and $P_{neg}$ is to affirm the personality description that aligns with the speaker $s$ and simultaneously negate the description that contradicts $s$'s personality. 

\subsection{Affective Dialog Content Construction}
\label{adc}
We first introduce how we construct the affective dialog content $X_{af}$ from the original dialog content $X$.

It is been pointed out that there are strong relations between the \textit{Extraversion} and \textit{Conscientiousness} traits and the positive affects such as \textit{Joy} and \textit{Surprise}; and between \textit{Neuroticism} and \textit{disagreeableness} and various negative affects such as \textit{Anger}, \textit{Sadness}, or \textit{Fear} \cite{watson1992traits}. By observing the dialogues in daily conversation, we also found that the utterance-level emotional interaction among the speakers (\textit{e.g.}, in what emotion the speaker $s$ talks to others or how others respond to $s$) is essential in analyzing the personality traits of $s$. 

Although affectivity is essential for personality recognition, in conversation scenarios, especially in real-time interaction between users and conversational agents, it is unrealistic to always have accurate affective annotations for utterances. Therefore, we fine-tuned a pre-trained language model $M_{ERC}$ especially for \textbf{E}motion \textbf{R}ecognition in \textbf{C}onversations to obtain the emotion label $e_i$ of each $u_i \in X$ in an offline manner:

\begin{equation}
	\small
	\begin{aligned}
	e_1, e_2, ..., e_m = M_{ERC}(u_1, u_2, ..., u_m)
	\end{aligned}
	\label{vad_personality}
\end{equation}
where each $e_i$ is one of the six basic emotions \cite{ekman1994nature} (\textit{i.e.}, \textit{Anger}, \textit{Disgust}, \textit{Fear}, \textit{Joy}, \textit{Sadness}, \textit{Surprise}) and \textit{Neutral}.
This emotion categories are widely employed in studies on personality analysis\cite{6376252,wen2021automatically}.

Although existing research \cite{broekens2023fine,amin2023wide} has validated the affective processing capabilities of language models for text, the interaction among speakers in conversational scenarios provides additional contextual cues for emotion recognition. Inspired by previous work on modeling dialogue flows \cite{gu2020dialogbert,kim2021emoberta,wolf2019huggingface}, we incorporate the speaker's identity into the utterance sequence during the encoding process to achieve more accurate emotion recognition, where the speaker's identity refers to binary indicators showing whether the current utterance is from the speaker $s$.


After we get each emotion label $e_i$, we wrap it into a handcraft template to form the affective description $a_i$ for the corresponding utterance $u_i$. The detailed process is depicted in Algorithm \ref{affective dialog content construction}, where the input consists of the dialog content accompanied by an emotion label sequence, and the output yields the dialog content enveloped by affective descriptions. We tried various templates and found that different template contents had little impact on the final effect, so we adopted the current illustrated template content.

\begin{algorithm}[t]
	\small
  \SetAlgoLined
  \KwData{Dialog Content $X=\{u_1, ..., u_m\}$, Speaker $s$, Other speakers $s_{other}$, Emotion labels $\{e_1, ..., e_m\}$}
  \KwResult{Affective Dialog Content $X_{af}$}
	$X_{af} \leftarrow $ ``"\;
  \For{$i\leftarrow 1$ \KwTo $m$}{
    \eIf{$i=1$}{ \tcp{ the first utterance}
    \eIf{$u_i$ is from $s$}{
      $a_i \leftarrow $  ``($s$ is initially $e_i$)" \;
      
            }{
      $a_i \leftarrow $  ``(At the beginning, $s_{other}$ is $e_i$)"
      }
    }{
      \eIf{$u_i$ is from $s$}{
      $a_i \leftarrow $  ``($s$ responds with $e_i$)"
            }{
      $a_i \leftarrow $  ``(Then, $s_{other}$ turns to be $e_i$)"
      }
    }
    $X_{af}$ append with $(u_i, a_i)$
  }
  \caption{\small{Affective Dialog Content Construction}}
  \label{affective dialog content construction}
\end{algorithm}

%

\subsection{Personality Description Construction}
\label{pd}
The class labels of the big five personality traits (such as Conscientiousness and Neuroticism) are not commonly used in everyday language. So, language models trained on general corpora may struggle to understand these terms, making personality identification challenging in conversational contexts.

The lexical hypothesis of personality traits states that (1) the most distinctive, significant, and widespread phenotypic attributes tend to become encoded as words in the conceptual reservoir of language \cite{goldberg1995so}. Therefore, we adopt and summarize descriptions of the big five personality traits in both positive and negative classes ($H_{pos}$, $H_{neg}$) from psychology findings \cite{saucier1996evidence,roccas2002big}, as shown in Figure \ref{personality_description}. These descriptions contain rich semantic meanings of personality traits in daily used languages that outline affective and behavioral tendencies of people with different personality traits. In Affective-NLI, these descriptions establish a correlation between the language and behavior of speakers during conversations and the personality traits, resulting in more understandable recognition results.

One possible concern is that these label descriptions may not be comprehensive or subjective in their wording. However, considering that these descriptions will be encoded in pre-trained models, benefiting from the powerful semantic understanding capabilities of language models, the semantics of each word within the descriptions, including its synonyms and similar expressions, will also be captured and encoded, to provide the auxiliary basis for personality recognition.

\begin{table*}[t]
    \centering
    \linespread{1.2}
    \small
    \caption{Positive and negative label descriptions of the Big Five personality traits}
    \begin{tabular}{m{2cm}m{2cm}p{12cm}}
        \toprule
        \textbf{Traits} & \textbf{Classes} & \textbf{Descriptions} \\
        \hline  
        \multirow{2}{*}{ \textbf{AGR}} & $H_{pos}$ & friendly, cooperative, empathetic, and compassionate, often prioritizing harmonious relationships and the well-being of others.\\
        & $H_{neg}$ & confrontational, uncooperative, lacking empathy, and often prioritizing their own needs and desires over the well-being of others. \\
        \hline 
        \multirow{2}{*}{ \textbf{CON}} & $H_{pos}$ & organized, responsible, diligent, detail-oriented, and committed to achieving their goals with a strong sense of duty and self-discipline.\\
        & $H_{neg}$ & disorganized, careless, impulsive, lacking discipline, and often displaying a disregard for responsibilities and commitments. \\
        \hline 
        \multirow{2}{*}{ \textbf{EXT}} & $H_{pos}$ & outgoing, sociable, energetic, and thriving in social interactions, often seeking stimulation and enjoying the company of others.\\
        & $H_{neg}$ & introverted, reserved, quiet, and often preferring solitude or smaller social settings, conserving energy and finding fulfillment in introspection and reflection. \\
        \hline 
        \multirow{2}{*}{ \textbf{NEU}} & $H_{pos}$ & prone to experiencing negative emotions, such as anxiety, worry, and mood swings, often displaying heightened sensitivity to stress and a tendency towards self-doubt and emotional instability.\\
        & $H_{neg}$ & emotionally stable, resilient, and composed, often displaying a calm and balanced demeanor, and having a tendency to handle stress and adversity with ease. \\
        \hline 
        \multirow{2}{*}{ \textbf{OPN}} & $H_{pos}$ & has curiosity, open-mindedness, creativity, tolerance, emotional expressiveness, and willingness to embrace new experiences and ideas.\\
        & $H_{neg}$ & closed-minded, resistant to change, lacking curiosity, intolerant of differences, emotionally guarded, and hesitant to explore new ideas or experiences. \\
        \bottomrule 
    \end{tabular}
    
    \label{personality_description}
\end{table*}

\subsection{NLI Training and Inference}
\label{ft}
To maximize the probability in Formula \ref{prob}, we construct NLI samples for fine-tuning the pre-trained language model $M$ in a prompt tuning manner.

We first wrap the original dialog content $X$ into the following two NLI input samples when recognizing each single personality trait $p$:

\begin{equation}
	\small
	\begin{aligned}
	\mathcal{T}_{pos}(X, p) &=  X_{af};H_{pos}(p);\texttt{Is it correct?[MASK].} \\
	\mathcal{T}_{neg}(X, p) &=  X_{af};H_{neg}(p);\texttt{Is it correct?[MASK].}
	\end{aligned}
\end{equation}
The model $M$ is fine-tuned to fill in the $\texttt{[MASK]}$ with ``yes'' and ``no''.

During training, we construct the NLI samples as follows:

\begin{equation}
	\small
	\begin{aligned}
	&\{ \mathcal{T}_{pos}(X, p), yes\}, \{ \mathcal{T}_{neg}(X, p), no\}; \ if \ y_p =1 \\
	&\{ \mathcal{T}_{pos}(X, p), no\}, \{ \mathcal{T}_{neg}(X, p), yes\}; \ if \ y_p =0 
	\end{aligned}
\end{equation}
where ``yes'' and ``no'' are the ground truth for NLI samples. According to the NLI samples, we denote the four probabilities as follows:

\begin{equation}
	\small
	\begin{aligned}
	& P_{pos}(yes) =  P(\texttt{[MASK]}=yes| \mathcal{M}_\theta(\mathcal{T}_{pos}(X, p))) \\
	& P_{pos}(no) =  P(\texttt{[MASK]}=no| \mathcal{M}_\theta(\mathcal{T}_{pos}(X, p))) \\
	& P_{neg}(yes) =  P(\texttt{[MASK]}=yes| \mathcal{M}_\theta(\mathcal{T}_{neg}(X, p))) \\
	& P_{neg}(no) =  P(\texttt{[MASK]}=no| \mathcal{M}_\theta(\mathcal{T}_{neg}(X, p))) \\
	\end{aligned}
\end{equation}

The training objective is fine-tuning the parameters $\theta$ in $\mathcal{M}$ to maximum the sum of the probabilities\:

\begin{equation}
	\small
	\begin{aligned}
	\mathop{\arg\max}\limits_{\theta} & (P_{pos}(yes)+ P_{neg}(no))*y_p \\ 
	 + & (P_{neg}(yes) + P_{pos}(no))*(1-y_p) 
	\end{aligned}
\end{equation}

In inference, the results are determined by the comparison of sums of the probabilities:

\begin{equation}
	\small
	\begin{aligned}
	\mathop{\arg\max} [P_{neg}(yes) + P_{pos}(no), P_{pos}(yes)+ P_{neg}(no)] 
	\end{aligned}
\end{equation}


%% file: experiment_new.tex
\section{Experiment Settings}
\subsection{Datasets}
Recording daily conversations for analysis, especially multi-person conversations, has potential privacy concerns. According to the book \textit{Television Dialogue: The sitcom Friends vs natural conversation} \cite{quaglio2009television}, television conversations and natural conversations are basically the same in terms of linguistic features. Therefore, we use two datasets analyzing the scripts of TV Series: FriendsPersona constructed by \cite{jiang2020automatic} and the Chinese Personalized and Emotional Dialogue (CPED) from \cite{chen2022cped} to evaluate Affective-NLI.

\textbf{FriendsPersona} is a dialog script dataset with personality annotations in 711 different dialogues, including 8,157 utterances. In each dialogue, The personality of one analyzed speaker is represented as 5-d binary vectors for the big-five traits. Following the problem settings in similar works \cite{rissola2019personality, jiang2020automatic} of personality analysis in conversation, we conduct binary classification tasks over the five personality traits respectively to evaluate our method. 

\textbf{CPED} is a large-scale Chinese personalized and emotional dialogue dataset, which contains around 12K dialogues of 392 speakers from 40 popular Chinese TV shows. To be consistent with the affective prompts and the personality descriptions in constructing input samples, we translate the dialogues in CPED into English with Google Translation\footnote{https://translate.google.com/} and manual verification. The detailed statistics are shown in Table \ref{datasets}.

\begin{table}[t]
    \centering
    \caption{Statistics of the two datasets.}
    \linespread{1}
    \renewcommand{\arraystretch}{1.2}
    \small
     \begin{tabular}{|l|c|c|}
        \hline
        \textbf{Statistics} & \textbf{FriendsPersona} & \textbf{CPED} \\
        \hline 
        \# Dialogues & 711 & 11,835\\
        \hline 
        \# Uttrs per dialogue  & 11.8 & 11.2 \\
        \hline 
        \# Unique Uttrs & 8,157 & 109,455 \\
        \hline 
        Uttr Length & 16.3 & 28.9 \\
        \hline 
        \multicolumn{1}{|p{2.5cm}|}{Label Distribution (Positive:Negative)} & \multicolumn{1}{m{2cm}|}{ AGR(.43:.57) CON(.46:.54) EXT(.44:.56) OPN(.35:.65) NEU(.47:.53)} & \multicolumn{1}{m{1.7cm}|}{AGR(.58:.42) CON(.67:.33) EXT(.65:.35) OPN(.50:.50) NEU(.59:.41)}\\
        \hline
    \end{tabular} 
    \label{datasets}
\end{table}

\begin{table*}[t]
    \centering
    \caption{Personality recognition accuracies on FriendsPersona and CPED.}
    \linespread{1}
    \renewcommand{\arraystretch}{1.2}
    \small
    \begin{tabular}{l|l|cccccc}
        \toprule
        \textbf{Dataset} & \textbf{Method} & \textbf{AGR} & \textbf{CON} & \textbf{EXT} & \textbf{OPN} & \textbf{NEU} & \textbf{Avg} \\
        \hline  
        \multirow{6}{*}{ \textbf{FriendsPersona}}
        & FT-RoBERTa (S) & 0.579$\pm$0.056 & 0.564$\pm$0.062 & 0.567$\pm$0.065 & 0.621$\pm$0.042 & 0.525$\pm$0.051 & 0.571\\
        & FT-RoBERTa (S+C) & 0.579$\pm$0.033 & 0.568$\pm$0.065 & 0.568$\pm$0.065 & 0.621$\pm$0.042 & 0.531$\pm$0.049 & 0.573\\
        & FT-RoBERTa (F) & 0.579$\pm$0.056 & 0.565$\pm$0.065 & 0.568$\pm$0.065 & 0.621$\pm$0.042 & 0.529$\pm$0.048 & 0.572 \\ 
        & DesPrompt  & 0.521$\pm$0.076 & 0.535$\pm$0.048 & 0.549$\pm$0.045 & 0.649$\pm$0.015 & 0.495$\pm$0.039 & 0.550\\  
        \cline{2-8}
        & Affective-NLI (RoBERTa) & 0.597$\pm$0.041 & 0.569$\pm$0.051 & 0.597$\pm$0.036 & 0.639$\pm$0.044 & \textbf{0.583}$\pm$0.069 & 0.597 \\
        & Affective-NLI (T5) & \textbf{0.625}$\pm$0.034 & \textbf{0.592}$\pm$0.039 & \textbf{0.614}$\pm$0.056 & \textbf{0.655}$\pm$0.076 & 0.581$\pm$0.073 & \textbf{0.613}\\ 
        & Affective-NLI (Llama) & 0.595$\pm$0.021 & 0.566$\pm$0.004 & 0.612$\pm$0.011 & 0.645$\pm$0.030 & 0.568$\pm$0.008 & 0.597 \\ 
        \hline  
        \multirow{6}{*}{ \textbf{CPED}}
        & FT-RoBERTa (S) & 0.602$\pm$0.012 & 0.654$\pm$0.011 & 0.660$\pm$0.017 & 0.529$\pm$0.012 & 0.595$\pm$0.014 & 0.608\\
        & FT-RoBERTa (S+C) & 0.599$\pm$0.016 & 0.656$\pm$0.012 & 0.659$\pm$0.017 & 0.529$\pm$0.014 & 0.600$\pm$0.016 & 0.608 \\
        & FT-RoBERTa (F) & 0.600$\pm$0.010 & 0.653$\pm$0.008 & 0.662$\pm$0.013 & 0.532$\pm$0.020 & 0.600$\pm$0.011 & 0.610 \\
        & DesPrompt  & 0.611$\pm$0.003 & 0.660$\pm$0.032 & 0.671$\pm$0.022 & 0.516$\pm$0.027 & 0.608$\pm$0.020 & 0.613\\
        \cline{2-8}
         & Affective-NLI (RoBERTa)  & 0.637$\pm$0.007 & 0.662$\pm$0.021 & 0.676$\pm$0.016 & 0.534$\pm$0.014 & 0.612$\pm$0.041 & 0.624\\ 
        & Affective-NLI (T5)  & \textbf{0.677}$\pm$0.014 & 0.656$\pm$0.013 & \textbf{0.689}$\pm$0.027 & \textbf{0.569}$\pm$0.018 & \textbf{0.647}$\pm$0.018 & \textbf{0.648} \\
        & Affective-NLI (Llama) & 0.645$\pm$0.033 &\textbf{0.668}$\pm$0.004 & 0.677$\pm$0.009 & 0.542$\pm$0.021 & 0.605$\pm$0.043 & 0.627\\ 
        \bottomrule 
    \end{tabular}
    \label{result_overall}
\end{table*}

\subsection{Methods in Comparison}
\label{baselines}

Affective-NLI mostly manipulates the input content, making it compatible with various large pre-trained language models (PLMs). Therefore, we implement it based on the three popular open-source PLMs in different architectures (\textit{i.e.}, encoder-only: RoBERTa \cite{liu2019roberta}, encoder-decoder: T5\cite{raffel2020exploring}, and decoder-only: Llama \cite{touvron2023llama}). We also compare Affective-NLI with two state-of-the-art methods \cite{jiang2020automatic,wen2023desprompt} in fine-tuning and prompt-tuning PLMs for personality recognition in conversation. Details of the methods are introduced as below.\\
\noindent
\textbf{Affective-NLI (RoBERTa)}: Here, we use the pre-trained Roberta-base model for Affective-NLI. RoBERTa \cite{liu2019roberta} is a famous pre-trained language model whose pre-training task is Masked-Language-Modeling (filling the masked tokens in sentences). It's natural to adopt it to conduct the mask-filling task in Affective-NLI.  \\
\noindent
\textbf{Affective-NLI (T5)}: Here, we use the pre-trained T5-base model as the backbone model of Affective-NLI. T5 \cite{raffel2020exploring} is a unified pre-trained language model to convert all text-based language problems into a sequence-2-sequence format. In affective-NLI, we use the token generated by the t5-decoder to fill in the \texttt{[MASK]} in affective-NLI. \\
\noindent
\textbf{Affective-NLI (Llama)}: Llama \cite{touvron2023llama} is one of the most popular large generative language models, comparable with ChatGPT\footnote{https://openai.com/blog/chatgpt} in some tasks\footnote{We didn't implement Affective-NLI on ChatGPT as it is not open-sourced.}. We adopt a parameter-efficient fine-tuning approach (\textit{i.e.} P-tuning \cite{liu2022p}) to only fine-tune a small number of extra parameters ($\sim$0.6\% of the whole model) while freezing most parameters of the pre-trained Llama-7b with around 7 billion parameters. Specifical to Affective-NLI, we use the last token generated by Llama to fill in the \texttt{[MASK]}.\\
\noindent
\textbf{FT-RoBERTa}: FT-RoBERTa\cite{jiang2020automatic} uses RoBERTa-base as the dialog encoder with three different kinds of input for personality recognition. FT-RoBERTa(S) only uses the utterances from the analyzed speaker $s$ as the input; FT-RoBERTa (S+C) concatenates utterances from the analyzed speaker and other utterances as context together as the input, while FT-RoBERTa (F) inputs all the utterances within the whole dialog flow in their natural order for classification. To better illustrate the three inputs: if a dialogue flow is $[U_1, C_1, U_2, U_3, C_2, ...]$, $U_i$ means the speaker's utterance and $C_i$ means the utterance from other speakers in the context. The input of FT-RoBERTa (S) is $[U_1, U_2, U_3, ...]$,  the FT-RoBERTa (S+C) input is $[U_1, U_2, U_3, ... , C_1, C_2, ...]$, while the FT-RoBERTa(F) input is $[U_1, C_1, U_2, U_3, C_2, ...]$ with the original sentence order.\\
\noindent
\textbf{DesPrompt}: DesPrompt \cite{wen2023desprompt} generates personality-descriptive prompts and fine-tunes pre-trained language models (\textit{i.e.}, here we use RoBERTa-base) for efficient personality recognition. It reformulates personality recognition as a cloze-style word-filling task and aligns with the pre-training Masked Language Modeling (MLM) task in most PLMs, hence enhancing the utilization of the pre-trained parameters. \\

\subsection{Evaluation Tasks and Metrics}
We split both datasets into train, validation, and test sets with a portion around 8:1:1. Models are trained on the train and validation sets and the results on test sets are reported.

To comprehensively evaluate the performance of Affective-NLI and other models, we design two personality recognition tasks in different settings specific to the conversation scenario: \textbf{Overall} and \textbf{Flow}. The \textbf{Overall} is the regular task that takes the full length of dialog content for personality recognition. The \textbf{Flow} examines the personality recognition performance at first 25\%, 50\%, 75\%, and the whole dialog flow. This is to show whether Affective-NLI can be applied to real conversation services to instantly recognize the personality of the user in the early stages of conversations. 

For both tasks, we use the binary classification accuracy to evaluate the personality recognition performance on each trait and the averaged result.

\subsection{Implementation Details}

During implementation, we pad all the utterances with [PAD] to the MAX\_LEN of 256. Each dialog flow is padded to a Dialogue MAX\_LEN of 20 according to the dataset statistics. The dialog flows are fed into the models in batches with a size of 32. As for the pre-trained Emotion Recognition in Conversation model $M_{ERC}$,  we classify 1000 randomly selected utterances in both datasets (in balanced emotion labels) by $M_{ERC}$. The accuracies are 0.792 for FriendsPersona and 0.815 for CPED, respectively, which is considerably accurate for the classification of seven classes. We use the Adam \cite{kingma2014adam} as the optimization algorithm in training. The learning rate for each model is selected to refer to the best performance on the validation sets.

%% file: result_analysis_new.tex
\section{Results Analysis}
Based on the aforementioned settings, we perform comprehensive experiments and conduct both quantitative and qualitative analyses of the obtained results. The quantitative analysis aims to address the following three Research Questions (\textbf{RQ}s), while the qualitative analysis utilizes a case study to showcase how Affective-NLI can provide interpretable personality recognition results.

\begin{itemize}
\item \textbf{RQ 1:} Can Affective-NLI outperform state-of-the-art approaches on personality recognition in conversation?
\item \textbf{RQ 2:} How much do different modules in Affective-NLI influence the performance in personality recognition?
\item \textbf{RQ 3:} How early can Affective-NLI recognize the personality in a dialog flow?
\end{itemize}

\begin{figure}[t]
\centering
\includegraphics[width=250pt,height=125pt]{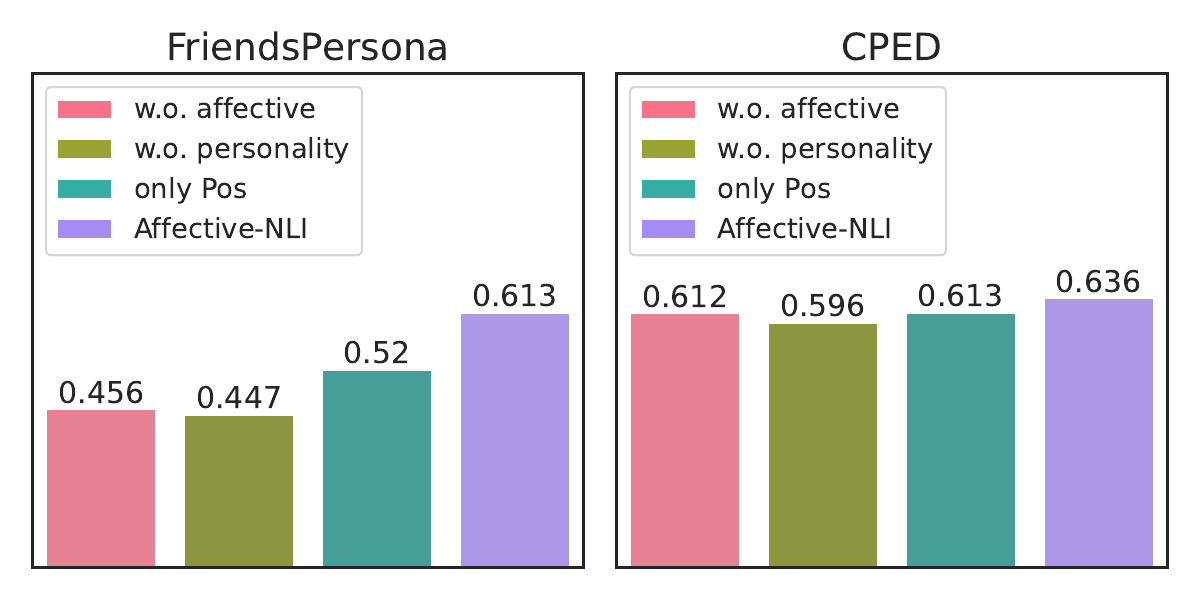}
\caption{Personality recognition accuracy comparison among the sub-models in the ablation study.}
\label{barplot}
\end{figure}

\subsection{Can Affective-NLI outperform state-of-the-art approaches on personality recognition in conversation (PRC)?}

To answer this research question, we analyze the PRC results of the methods introduced in Section \ref{baselines} on FriendsPersona and CPED are shown in Table \ref{result_overall}. These results are obtained from the \textbf{Overall} experiment setting.

On both datasets, Affective NLI achieves the best accuracies (including single personality trait recognition and overall performances) compared with other methods, which validates its effectiveness in personality recognition in conversations. Specifically,  the best Affective-NLI method (\textit{i.e.}, Affective-NLI (T5)) outperforms the best baseline methods on both datasets by around 6-7\%.



Due to different distributions, the models' performance in recognizing the same personality trait varies across the two datasets. However, we can still observe that Affective-NLI exhibits the highest improvement over other methods in terms of AGR and NEU. We hypothesize that this may be attributed to a greater disparity in affective information expressed by positive and negative speakers in conversations pertaining to these two personality types.


Within the three implementations, Affective-NLI (T5) achieved the highest accuracy in recognizing seven out of the ten personality traits. This resulted in Affective-NLI (T5) having the best average results on both datasets. This suggests that compared to other types of pre-trained language models, the encoder-decoder architecture of the T5 model performs better in this form of NLI task. In addition, Affective-NLI (RoBERTa) and Affective-NLI (Llama) performed similarly, indicating that the parameter size of the pre-trained models does not play a decisive role in the results of the current PRC task. Therefore, Affective-NLI can be implemented on slightly smaller pre-trained language models without sacrificing accuracy, which validates the usability of Affective-NLI.

Although our work has achieved state-of-the-art performance compared to most existing models, we are aware that the current performance still has some distance from real-world applications. Addressing this issue more effectively will require further efforts from interdisciplinary work in the future.

\begin{table*}[t]
    \centering
    \caption{Dialogue Samples for Personality Recognition.}
    \linespread{1.1}
    \renewcommand{\arraystretch}{1.2}
    \small
    \resizebox{\linewidth}{!}{
    \begin{tabular}{p{10cm}ccccc}
		\hline
        \multirow{2}{*}{\textbf{Dialog Content}} &  \multicolumn{5}{c}{\textbf{Probability of Personality}} \\ 
        \cline{2-6}
        & \textbf{AGR} & \textbf{CON} & \textbf{EXT} & \textbf{OPN} & \textbf{NEU}  \\
		\hline
        \textbf{Agent:} Good morning, Mrs. Thompson! How are you feeling today? \\ (In the beginning, the agent is in \textbf{Joy}) & \multicolumn{5}{c}{-}\\
        \textbf{Mrs. Thompson:}  Oh, \textbf{everything is falling apart!} My arthritis is acting up, my cat ran away, and the weather outside is simply \textbf{dreadful}! \\ (Mrs. Thompson responds with \textbf{Anger}) & \makecell*[c]{37\%} & 42\% & 45\% & 36\% & \textbf{54\%} \\  
        \textbf{Agent:}  I understand, Mrs. Thompson. How can I help you? \\ (Then, the agent turns to \textbf{Neutral}) & \multicolumn{5}{c}{-}\\     
        \textbf{Mrs. Thompson:}   I don't know. It just feels like \textbf{the world is against me}... I \textbf{can't seem to catch a break}... \\ (Mrs. Thompson responds with \textbf{Sadness})
& 21\% & 41\% & 47\% & 28\% & \textbf{68\%} \\ 
		 \textbf{Agent:} It's understandable to feel overwhelmed.   Have you ever tried engaging in relaxation   exercises or practicing mindfulness during   challenging moments? & \multicolumn{5}{c}{-}\\
       	\textbf{...} & \multicolumn{5}{c}{...}\\ 
		\hline
    \end{tabular}
    }
    \label{case_study}
\end{table*}

\subsection{How much do different modules in Affective-NLI influence the performance in personality recognition?}

To figure out the answer, we design an ablation study to show the results of sub-models without different modules in Affective-NLI, respectively.
We selected the Affective-NLI (T5) model with the best overall performance and designed the following sub-models:

\noindent
\textbf{w.o. affective}: we remove the affective annotations, replacing $X_{af}$ with the original $X$. However, we retained the personality description in the input. This is to evaluate the effect of affective information on personality recognition.

\noindent
\textbf{w.o. personality}: we keep the affective annotations and the NLI task format in the input but replace the personality label descriptions as directly the personality label name. This is to investigate the influence of semantic information contained in the personality descriptions on the PRC task.

\noindent
\textbf{only Pos}: in this sub-model, we maintained the input of Affective-NLI but only used positive personality descriptions during training and inference. This is to examine whether combining two different descriptions of the same personality enhances performance.

We conduct the personality recognition in conversation on FriendsPersona and CPED with all the sub-models, their accuracies are shown in Figure \ref{barplot}. These results are also obtained from the \textbf{Overall} experiment setting.

In general, removing any module in Affective-NLI results in a decrease in model performance,  which validates our model design. Specifically, when removing the personality description (\textbf{w.o. personality}), the performance of Affective-NLI drops significantly on both datasets, highlighting the importance of enabling the model to understand the semantics of personality for personality recognition. Besides, removing the affective annotations (\textbf{w.o. affective}), also weakens the model to a considerable extent, particularly evident in the FriendsPersona dataset. By referring to Table \ref{datasets}, it can be observed that utterance length in FriendsPersona is apparently shorter. This suggests that providing affective information can be more helpful for personality recognition when there is less conversation content available. When using only the positive descriptions of each personality trait in Affective-NLI (\textbf{only Pos}), the model also shows a certain degree of performance decline, with a more pronounced decrease in FriendsPersona. Therefore, supervising the model to learn semantic information from different categories of personality traits can provide auxiliary benefits.

\begin{figure}[t]
\centering
\includegraphics[width=250pt,height=110pt]{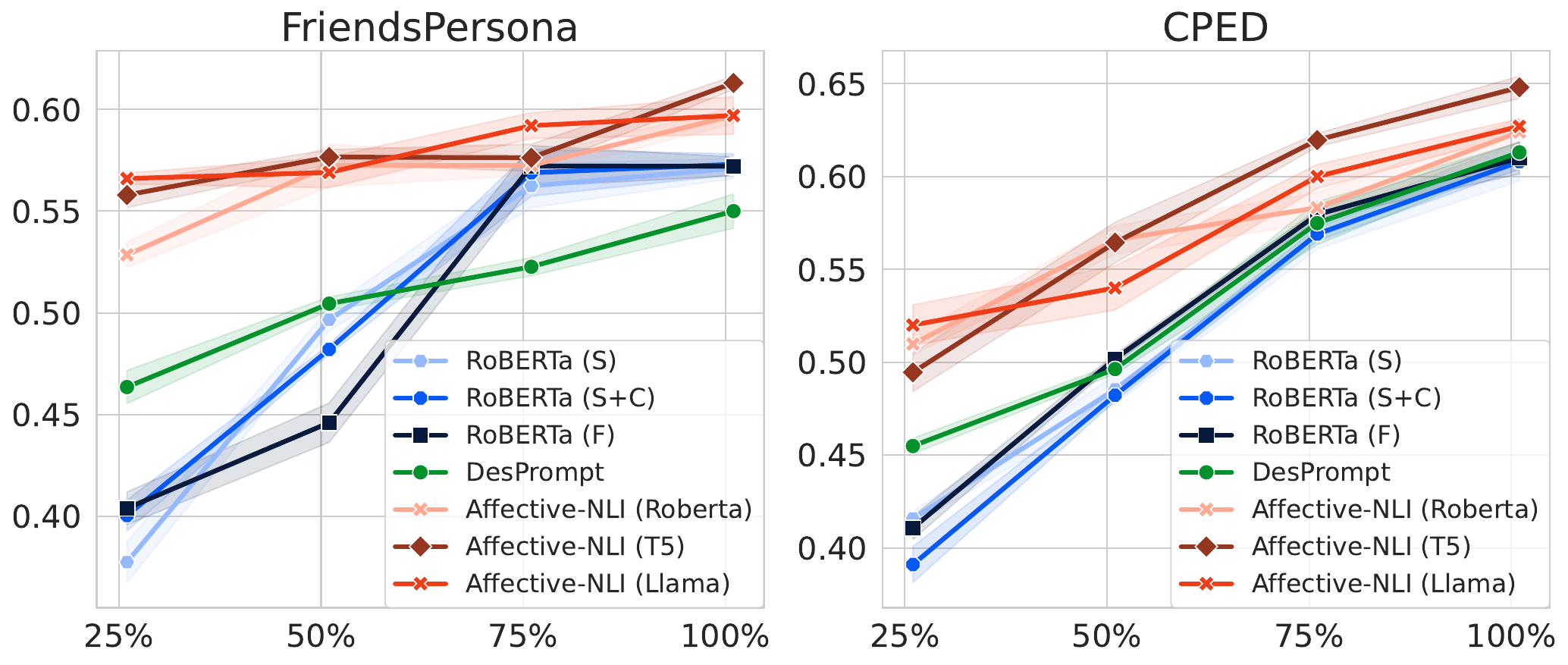}
\caption{The personality recognition accuracy comparison in the \textbf{Flow} experiment. The transparent error bars indicate the lower and upper bounds of the current accuracies.}
\label{dialog_flow}
\end{figure}

\subsection{How early can Affective-NLI recognize the personality in a dialog flow?}

To evaluate whether Affective-NLI can be applied to conversation scenarios for instant personality recognition, we design an experiment \textbf{Flow} by feeding the first 25\%, 50\%, 75\%, and 100\% of each dialog flow in Test sets to the personality recognition models. According to the different lengths of dialog flows, we calculate the average numbers of utterances from the analyzed speaker $s$ are 0.52, 1.48, 2.63, and 4.09, respectively. So, we make an additional rule that each dialog flow fed into the models must contain at least one utterance from the analyzed speaker $s$. Based on the aforementioned data settings, the average accuracy of all models for personality recognition of the five personality traits on both datasets is shown in Figure \ref{dialog_flow}.

When the dialog flow is at 25\%,  Affective-NLI models
significantly outperform other methods. It suggests that in the early stage of a dialogue (only one of two sentences from the analyzed speaker), the affective information and label semantics provide additional evidence to analyze the personality effectively. DesPrompt also performs relatively well as it provides an ample amount of descriptive label words.

When feeding 50\%-75\% of the whole dialog flow, Affective-NLI models also outperform the rest of the models. However, fine-tuning RoBERTa models (FT-RoBERTa (S), FT-RoBERTa (S+C), and FT-RoBERTa (F)) improve dramatically, and even outperform DesPrompt on FriendsPersona. This demonstrates the importance of rich semantic content in dialogues for traditional classification methods. Therefore, Affective-NLI effectively addresses the challenge of insufficient semantic information in early-stage conversations, which is a limitation of traditional classification methods.

The improvement rate of the model's performance slows down from feeding 75\% to the entire dialog flow. This suggests that the relationship between semantic information and effectiveness in personality recognition during a conversation is not simply linearly correlated.

To summarize, the strength of integrating affective information and label semantics in Affective-NLI is reflected in the early stage of conversations. With only one or two utterances from the analyzed speaker, Affective-NLI can instantly recognize the personality with around 0.5-0.6 in accuracy.

\subsection{Case Study}
We use a case study to illustrate the usage of Affective-NLI in real HCI application scenarios, as shown in Table \ref{case_study}.

Affective-NLI is designed for personality recognition in conversations. Therefore, the main experiments intend to verify the personality recognition performance with existing dialog pieces. However, in real HCI applications, we still need to design conversation agents to select appropriate, emotional, and personalized dialogue strategies based on the recognized personality. The case study here demonstrates how Affective-NLI supports such dialogue strategies during interactions. In this example, we pretend to be Mrs. Thompson and engage in a conversation with the Agent. The probability of personality in the case study is obtained from Affective-NLI, and the Agent's responses are generated by ChatGPT based on the identification results from Affective-NLI.

In Table \ref{case_study}, the dialog content with affective annotations is sequentially inputted into Affective-NLI to recognize the personality trait of Mrs. Thompson. Specifically, when Mrs. Thompson responds to the Agent, Affective-NLI calculates the probabilities (confidences of recognition results) of Mrs. Thompson exhibiting all the big five personality traits displayed on the right side of the Table. To facilitate decision-making, we can set a probability threshold, denoted as $\delta$ (\textit{e.g.}, \textbf{60\%}). When the probability of Mrs. Thompson possessing a certain personality trait (1) exceeds $\delta$ and (2) is the highest among the five personality traits, the Agent can generate a response with a specific dialogue policy tailored to that personality trait.

%% file: limitation.tex
\section{Discussion}
In this section, we discuss three practical issues of utilizing Affective-NLI in real applications.

Firstly, the recognition results of Affective-NLI cannot be directly used as diagnostic evidence for psychological therapy. Personality is a multifaceted concept that is investigated and discussed in various disciplines, including physiology, linguistics, and cognitive science. Therefore, rigorous personality assessment still relies on professional approaches. The purpose and advantage of Affective-NLI lie in its ability to provide preliminary evaluations of users' personalities in conversational settings, facilitating a better conversational experience.

Secondly, Affective-NLI can only recognize the personality manifested during conversations, which may not reflect the users' actual personalities. The content of conversations is constrained by different contexts, and individuals may not fully express their personalities due to politeness, cultural factors, and other considerations in specific situations. However, this limitation is inherent to the problem of personality recognition in conversation itself.

Thirdly, both datasets used to evaluate Affective-NLI consist of scripts from TV series rather than real-life conversations. Collecting conversations for personality analysis should be cautiously approached due to potential ethical and privacy concerns. To provide a stronger rationale for future data collection and usage, we initially validate our concept using dialog scripts from TV series. However, since the scripts we use are derived from TV series in various cultures and contexts, our model can still be either directly applied or fine-tuned with a small amount of real data to adapt to real application scenarios.